\documentclass[11pt,a4paper]{article}
\usepackage[hyperref]{naaclhlt2018}
\usepackage{times}
\usepackage{latexsym}
\usepackage{graphicx}
\usepackage{url}
\usepackage{enumitem}
\usepackage{amsmath}
\usepackage{amsfonts}
\usepackage{nth}
\usepackage{fancyvrb,cprotect}

\aclfinalcopy % Uncomment this line for all SemEval submissions

\title{AttnConvnet at SemEval-2018 Task 1:  Attention-based Convolutional Neural Networks for Multi-label Emotion Classification}
\author{Yanghoon Kim\textsuperscript{1,2}, Hwanhee Lee\textsuperscript{1} and Kyomin Jung\textsuperscript{1,2} \\
  \textsuperscript{1}Seoul National University, Seoul, Korea \\
  \textsuperscript{2}Automation and Systems Research Institute, Seoul National University, Seoul, Korea\\
  {\tt \{ad26kr,wanted1007,kjung\}@snu.ac.kr} \\
}  
\date{}

\begin{document}
\maketitle
\begin{abstract}
In this paper, we propose an attention-based classifier that predicts multiple emotions of a given sentence. Our model imitates human's two-step procedure of sentence understanding and it can effectively represent and classify sentences. With emoji-to-meaning preprocessing and extra lexicon utilization, we further improve the model performance. We train and evaluate our model with data provided by SemEval-2018 task 1-5, each sentence of which has several labels among 11 given emotions. Our model achieves \nth{5}/\nth{1} rank in English/Spanish respectively.
% * <kindaichi7207@gmail.com> 2018-02-22T02:00:06.018Z:
% 
% human mechanism이라고 하는데 이부분이 multiple emotion 을 classify 하는 부분이라는 설명이 필요할듯 / redundant weights 부분은 빼는게 좋지 않을까?(저 부분이 나오면 논문에 왜 redundant 한지 설명이 필요할 것 같음)
% 
% ^ <ad26kt@gmail.com> 2018-02-22T12:13:21.542Z:
% 
% 앱스트랙트 아직 안고침 ㅋ
%
% ^.
\end{abstract}

\section{Introduction}

Since the revolution in deep neural networks, especially with the help of Long short-term memory\cite{lstm}, it has been easy for machines to imitate human's linguistic activities, such as sentence classification\cite{yoon}, language model\cite{lm}, machine translation\cite{Bahdanau}.
% * <kindaichi7207@gmail.com> 2018-02-22T02:02:13.870Z:
% 
% Since -> Due to 가 맞을듯 (Since는 접속사)
% 
% ^ <ad26kt@gmail.com> 2018-02-22T12:13:33.582Z:
% 
% 전치사로써 쓰이면 맞는데?
%
% ^.

Emotion classification is a subpart of sentence classification that predicts the emotion of the given sentence by understanding the meaning of it. Multi-label emotion classification requires more powerful ability to comprehend the sentence in variety of aspects. For example, given a sentence 'For real? Look what I got for my birthday present!!', it is easy for human to figure out that the sentence not only expressing 'joy' but also 'surprise'. However, machines may require more task-specific structure to solve the same problem.

Attention mechanisms are one of the most spotlighted trends in deep learning and recently made their way into NLP. Applied to systems with neural networks, it functions as visual attention mechanisms found in humans\cite{attention} and the most effective region of features will be highlighted over time, making the system better exploit the features related to the training objective. \cite{Bahdanau} is one of the most significant footprints of attention mechanism in NLP and they applied attention mechanisms to machine translation for the first time. The model generates target word under the influence of related source words. Furthermore, \citeauthor{transformer} \shortcite{transformer} proposed a brand new architecture for neural machine translation. The model utilizes attention mechanisms not only as the submodule but also as the main structure, improving time complexity and performance.

Inspired by \cite{transformer}, we come up with attention-based multi-label sentence classifier that can effectively represent and classify sentences. Our system is composed of a self-attention module and multiple CNNs enabling it to imitate human's two-step procedure of analyzing sentences: comprehend and classify. Furthermore, our emoji-to-meaning preprocessing and extra lexicon utilization improve model performance on given dataset. We evaluated our system on the dataset of \cite{SemEval2018Task1}, where it ranked \nth{5}/\nth{1} rank in English/Spanish respectively.
% * <kindaichi7207@gmail.com> 2018-02-22T02:04:39.515Z:
% 
% > that it -> that
% ? 이미 수정한건가?
% 
% ^.

%The rest of the paper is structured as follows. Section
%2 discusses in brief the dataset for the task.
%Section 3 explains the various approaches used by
%our ensemble model, the kind of experiments we
%carried out along with the details of the parameters
%which gave optimal results on cross validation,
%and the way we combined the predictions.
%Section 4 explains how the system is evaluated
%and Section 5 states the results we achieved and
%discusses the various implications of those results.
%We conclude our work in Section 6.
\begin{figure*}[ht]
	\centering
  	\includegraphics[width = \textwidth ]{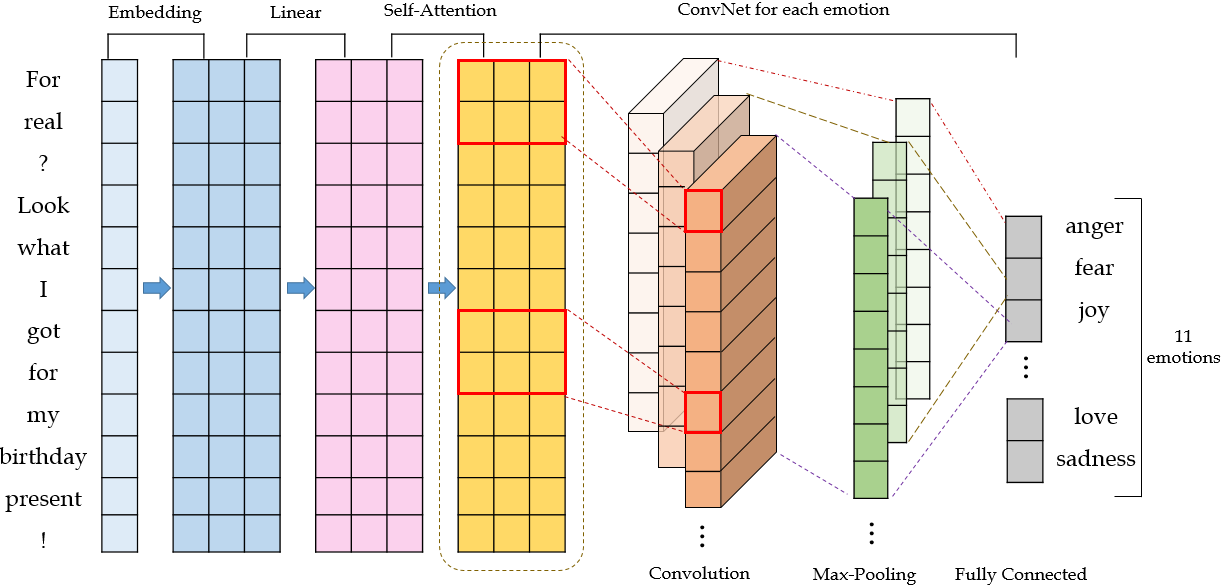}
	\caption{Overall architecture of the model. Preprocessed data goes through embedding layer, self-attention layer, Convolution layer and pooling layer step by step.}
  	\label{fig:architecture}
\end{figure*}

\section{Model}
Our system is mainly composed of two parts: self-attention module and multiple independent CNNs as depicted in Figure \ref{fig:architecture}. This structure is actually imitating how human perform the same task. In general, human firstly read a sentence and try to comprehend the meaning, which corresponds to self-attention in our system. Then human categorize the sentence to each emotion separately but not all at once, and that is the reason why our system use 11 independent CNNs. In addition to main structure, we added the description of preprocessing in the model description because it makes up a large proportion in NLP tasks, especially when the dataset is small. Details are described in the following paragraph step by step.

{\bf Preprocessing}:
For raw data, we applied 3 steps of preprocessing: 
\begin{enumerate}[label=(\roman*)]
% * <kindaichi7207@gmail.com> 2018-02-21T08:00:30.253Z:
% 
% In consideration 문장은 별도로 앞에 빼는게 좋을듯 emoji 부분이 너무 길다는 생각
% 
% ^ <ad26kt@gmail.com> 2018-02-21T14:19:01.019Z.
\item Our system mainly deals with limited numbers of tweet data, which is very noisy. In this case, preprocessing of data has crucial impact on model performance. Emoji may be referred to as a typical property of tweets and we found that considerable number of tweets contain emojis. Each emoji has a meaning of their own, and we converted every emoji in the data to phrase/word that represents its meaning. We call this procedure as \textbf{emoji-to-meaning} preprocessing. Some tweets have too many repetition of certain emoji that may make the sentence over-biased to certain emotions. Against expectations, removing overlapped emojis reduced performance. 
% * <kindaichi7207@gmail.com> 2018-02-22T02:08:02.062Z:
% 
% >  especially validation data and test data. 이 부분은 validation data랑 test data를 보고 끼워 맞췄다는 인상을 줄 수 있어서 빼는게 좋을 것 같음
% 수정했음, 좀 맘에 안들기는 한데 더 좋은 문장 있으면 알려줘
% 
% ^.
\item Lower-case and tokenize data with \textbf{TweetTokenizer} in \cite{nltk}.
\item Remove all of the mentions and '\#' symbols in the beginning of all topics. Unlike mentions, topics may include emotional words and hence we don't remove the topic itself.
% * <kindaichi7207@gmail.com> 2018-02-22T02:12:10.675Z:
% 
% > that -> so나 hence같이 그래서가 어감이 맞는듯
% ok
% 
% ^.
\end{enumerate}

{\bf Embedding}: It is especially helpful to use pre-trained word embeddings when dealing with a small dataset. Among those well-known word embeddings such as Word2Vec\cite{word2vec}, GloVe\cite{glove} and fastText\cite{fasttext}, we adopt 300-dimension GloVe vectors for English ,which is trained on Common Crawl data of 840 billion tokens and 300-dimension fastText vectors for Spanish, which is trained on Wikipedia.
% * <kindaichi7207@gmail.com> 2018-02-22T02:17:00.263Z:
% 
% > for Enlgish, trained with Common Crawl data of 840 billion tokens, fastText for Spanish which has 300-dimension trained on WikiPedia. 스페인어 추가함
% ok
% 
% ^.

\begin{figure}
	\centering
	\includegraphics[width = 0.45\textwidth]{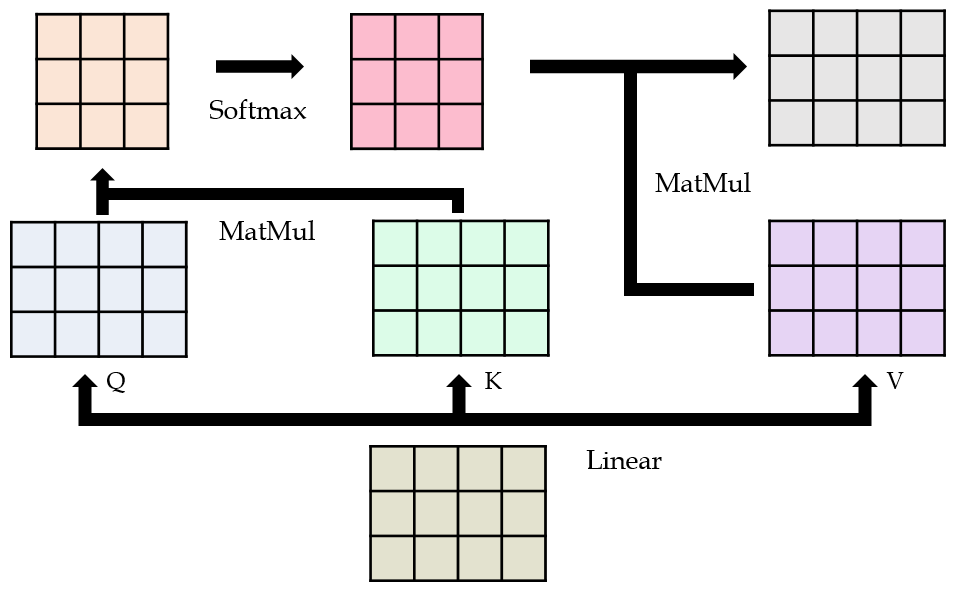}
  	\caption{Inner architecture of self-attention module}
    \label{fig:self-attention}
\end{figure}

{\bf Self-attention}: \citeauthor{transformer} \shortcite{transformer} proposed a non-recurrent machine translation architecture called Transformer that is based on dot-product attention module. Usually, attention mechanisms are used as a submodule of deep learning models, calculating the importance weight of each position given a sequence. In our system, we adopt the self-attention mechanisms in \cite{transformer} to represent sentences. The detailed structure of self-attention is shown in Figure \ref{fig:self-attention}. Dot-product of every embedded vector and weight matrix \(W \in \mathbb{R}^{d_e\times 3d_e}\) is split through dimension as \(Q\), \(K\), \(V\) of the same size, where \(d_e\) is the dimensionality of embedded vectors. Then attended vector is computed as in \eqref{eq_3}. 

\begin{gather}
E = [emb(x_1), emb(x_2), ..., emb(x_n)] \label{eq_1} \\
[Q, K, V] = [eW \; for \; e \; in \; E] \label{eq_2} \\
Attn(Q, K, V) = softmax(\frac{QK^T}{\sqrt[]{d_e}})V \label{eq_3}
\end{gather}

Multi-head attention allows the model to benefit from ensemble effect only with the same amount of parameter.
\begin{gather}
Multihead(Q, K, V) = Concat(head_1, ..., head_h) \\
\text{where} \; head_i = Attn(Q_i, K_i, V_i) \nonumber \\
Q = [Q_1, ..., Q_h], \; Q_i \in \mathbb{R}^{n \times \frac{d_e}{h}} \nonumber \\
K = [K_1, ..., K_h], \; K_i \in \mathbb{R}^{n \times \frac{d_e}{h}} \nonumber \\
V = [V_1, ..., V_h], \; V_i \in \mathbb{R}^{n \times \frac{d_e}{h}} \nonumber
\end{gather}
For each self-attention layer, there are additional position-wise feed-forward networks right after the attention submodule.
\begin{gather}
FFN(x) = max(0, xW_1 + b_1)W_2 + b_2 \\
\text{where} \; W_1 \in \mathbb{R}^{d_e \times d_f}, \; W_2 \in \mathbb{R}^{d_f \times d_e}
\end{gather}
In addition to these two sub-layers, there is a residual connection around each sub-layer, followed by layer normalization. Also, we can stack self-attention layers by substituting the embedding vectors in \eqref{eq_1} with the output of the last self-attention layer.

{\bf Convolution \& Pooling}: Followed by self-attention layer are 11 independent 1-layer Convolution layers with max-pooling layers. \citeauthor{yoon} \shortcite{yoon} has proved that CNNs have lots of potential in sentence processing task and we adopt the CNNs in the same way.
% * <kindaichi7207@gmail.com> 2018-02-22T02:18:49.556Z:
% 
% > adopt -> adopted
% nope
% 
% ^.

{\bf Output \& Loss}: Each output of CNNs go through a fully-connected layer to generate a logit. Sigmoid activation is applied to calculate the probability of each emotion, and we use the sum of each class' cross-entropy as the final loss function.
% * <kindaichi7207@gmail.com> 2018-02-22T02:19:08.705Z:
% 
% > a  -> 삭제
% nope
% 
% ^.

% Min: no longer used as of NAACL-HLT 2018, following ACL exec's decision to
% remove this extra workflow that was not executed much.
% BEGIN: remove
%% \section{XML conversion and supported \LaTeX\ packages}

%% Following ACL 2014 we will also we will attempt to automatically convert 
%% your \LaTeX\ source files to publish papers in machine-readable 
%% XML with semantic markup in the ACL Anthology, in addition to the 
%% traditional PDF format.  This will allow us to create, over the next 
%% few years, a growing corpus of scientific text for our own future research, 
%% and picks up on recent initiatives on converting ACL papers from earlier 
%% years to XML. 

%% We encourage you to submit a ZIP file of your \LaTeX\ sources along
%% with the camera-ready version of your paper. We will then convert them
%% to XML automatically, using the LaTeXML tool
%% (\url{http://dlmf.nist.gov/LaTeXML}). LaTeXML has \emph{bindings} for
%% a number of \LaTeX\ packages, including the NAACL-HLT 2018 stylefile. These
%% bindings allow LaTeXML to render the commands from these packages
%% correctly in XML. For best results, we encourage you to use the
%% packages that are officially supported by LaTeXML, listed at
%% \url{http://dlmf.nist.gov/LaTeXML/manual/included.bindings}
% END: remove

\section{Experiments \& Results}

\subsection{Data}
For the SemEval 2018 shared task, \citeauthor{SemEval2018Task1}\shortcite{SemEval2018Task1} has provided tweet data with multiple labels among 11 pre-set emotions: 'angry', 'anticipation', 'disgust', 'fear', 'joy', 'love' 'optimism', 'pessimism', 'sadness', 'surprise' and 'trust'. We only use English and Spanish data among three different languages. The dataset consists of 6838/887/3259 tweets in English, 3561/679/2854 tweets in Spanish for train/validation/test data respectively.

\subsection{Setup}
We implemented a model with 3-layer self-attention and 1-layer CNN. With the restriction of fixed-size GloVe vector, we found that 300-dimension hidden state is excessive for such a small dataset that we added a position-wise linear layer between the embedding layer and self-attention layers to make \(d_e = 30\). We employed \(h = 2\) for multi-head attention and set \(d_f = 64\). Two regularization techniques are applied to our system: Dropout with \(P_{drop} = 0.1\) for self-attention, and L2 regularization for all weight matrix but not bias. We added 0.001 times regularization loss to original loss function. We optimized the loss with Gradient Descent using Adam optimization algorithm with additional learning rate decay.

\subsection{Model variants}
\label{model variants}
% * <kindaichi7207@gmail.com> 2018-02-22T02:20:15.947Z:
% 
% +, - 보다 with without(wO) 형태로 많이 쓰는듯 
% 윗부분 제대로 안읽으면 AC - attn 이 AC에 attn을 뺀게 아니라 추가한걸로 착각할수도 있음
% 
% ^ <ad26kt@gmail.com> 2018-02-22T12:38:12.392Z:
% 
% 응 근데 기본적으로 읽었다는걸 가정해야 맞는거 아닌가? ac w/o 으로 하면, 뭔가 ac의 특성이 남아있어야 할것같은데, 사실 attn 빼버리면 그냥 cnn이니까 그게 좀 마음에 걸려서 일부러 저렇게 썼는데
%
% ^.
We conduct experiments with following variants of our model.
\begin{itemize}
\item \textbf{AC}:
Self-attention + CNNs, which is our basic system.

\item \textbf{AC - attn}:
Basic system without self-attention module.

\item \textbf{AC + nrc1}:
We mainly used NRC Emotion lexicon\cite{nrc} to make word-level label of each sentence, counting the occurence of each emotion in the sentence. Each of the word-level label is concatenated to the output vector of each pooling layer.

\item \textbf{AC + nrc2}:
At evaluation/test step, binarize the word-level label and add 0.4 times the label value to the logit. 

\item \textbf{AC + synth}:
Inspired by \cite{monolingual}, we made synthetic data using unlabeled SemEval-2018 AIT DISC data\footnote{https://www.dropbox.com/s/2phcvj300lcdnpl/SemEval2018-AIT-DISC.zip?dl=0} with pre-trained model, and fine-tuned the model with synthetic data.

\end{itemize}
\subsection{Experimental results}
We conduct several experiments to prove the effectiveness of our model, each to verify the benefit from: (1) tweets specific preprocessing (2) self-attention representation (3) emotional lexicon utilization. Experimental results are \textbf{mainly compared with English data.}

\subsubsection{Impact of emoji-to-meaning}
We firstly verify the efficiency of emoji-to-meaning preprocessing. Table \ref{table:1} shows the accuracies of the same model with different preprocessing. We found that emoji-to-meaning preprocessing can improve the model accuracy by 1\%. When a emoji is converted to its meaning, it can be represented as a combination of emotional words allowing it to not only reduce redundant vocabulary but also further emphasize the influence of certain emotions.
% * <kindaichi7207@gmail.com> 2018-02-22T02:22:20.741Z:
% 
% > find -> found
% ok
% 
% ^.
% * <kindaichi7207@gmail.com> 2018-02-22T02:21:52.054Z:
% 
% > efficacy -> efficiency
% ok
% 
% ^.
% * <kindaichi7207@gmail.com> 2018-02-22T02:21:33.880Z:
% 
% > verify -> verified
% nope
% 
% ^.

\begin{table}[h!]
\begin{tabular}{ |p{2cm}||p{2.5cm}|p{2.3cm}| }
\hline
Model & Accuracy(valid) & Accuracy(test)\\
\hline
AC	(w/o)&54.86\%	&54.91\%  \\
\hline
AC 	&55.94\%	&55.90\%  \\
\hline
\end{tabular}
\caption{Experimental results with and without emoji-to-meaning preprocessing.}
\label{table:1}
\end{table}

\subsubsection{Impact of self-attention}
To examine the effectiveness of self-attention representation, we simply get rid of self-attention layers. Table \ref{table:2} shows that by removing the self-attention layers, both the validation/test accuracy dropped over 4\%. This may be attributed to the ability of self-attention: It helps the model to better learn the long-range dependency of sentences. Learning long-range dependencies is a key challenge in NLP tasks and self-attention module can shorten the length of paths forward and backward signals have to traverse in the network as described in \cite{transformer}.
% * <kindaichi7207@gmail.com> 2018-02-22T02:26:04.491Z:
% 
% > get -> got
% method 관련 된건 현재형으로 표현
% 
% ^.
\begin{table}[h!]
\begin{tabular}{ |p{2cm}||p{2.5cm}|p{2.3cm}| }
\hline
Model & Accuracy(valid) & Accuracy(test)\\
\hline
AC - attn	&51.04\%	&51.60\%  \\
\hline
AC 	&55.94\%	&55.90\% \\
\hline
\end{tabular}
\caption{Comparison between our basic system and basic system without self-attention module.}
\label{table:2}
\end{table}

\subsubsection{{Impact of extra resources}}
Lack of data has crucial impact on model generalization. Generalization techniques such as dropout or L2 regularization can relieve over-fitting problem to a certain extent; however, it can't totally substitute the effect of rich data. So we apply some heuristic methods to exploit extra resources as described in \ref{model variants}. Table \ref{table:2} shows that model can slightly benefit from extra lexicon if used properly. However, adding synthetic data which is made from pre-trained model didn't help a lot, and in some cases even reduce the accuracy of the test result. Actually, \citeauthor{monolingual}\shortcite{monolingual} emphasized that they used the monolingual sentences as the target sentences, informing that the target-side information, which corresponds to label in our task, is not synthetic. However, we made synthetic labels with a pre-trained model and it may only cause over-fitting problem to the original training data.
\begin{table}[h!]
\begin{tabular}{ |p{2cm}||p{2.5cm}|p{2.3cm}| }
\hline
Model & Accuracy(valid) & Accuracy(test)\\
\hline
AC	&55.94\%	&55.90\%  \\
\hline
AC + nrc1	&56.13\%	&56.02\% \\
\hline
AC + nrc2	&57.16\%	&56.40\% \\
\hline
AC + synth	&55.88\%	&55.90\% \\
\hline
\hline
Ensemble	&\textbf{59.76}\%	&\textbf{57.40}\% \\
\hline
\end{tabular}
\caption{Experimental results with extra resources and an ensemble result}
\label{table:3}
\end{table}

\subsubsection{Ensemble}
Our best results are obtained with an ensemble of 9 parameter sets of AC + nrc2 model that differ in their random initializations. The ensemble model achieved validation/test accuracy of 59.76\%/57.40\% in English data and 50.00\%/46.90\% in Spanish data respectively.

\section{Conclusion}
\label{sec:length}
In this paper, we proposed an attention-based sentence classifier that can classify a sentence into multiple emotions. Experimental results demonstrated that our system has effective structure for sentence understanding. Our system shallowly follows human's procedure of classifying sentences into multiple labels. However, some emotions may have some relatedness while our model treats them independently. In our future work, we would like to further take those latent relation among emotions into account.
% * <kindaichi7207@gmail.com> 2018-02-22T02:25:31.310Z:
% 
% > classifying sentences -> classifying multiple labeled sentences ?
% ok
% 
% 
% ^.
% * <kindaichi7207@gmail.com> 2018-02-22T02:25:23.750Z:
%
% ^.

\section*{Acknowledgments}
This work was supported by the National Research Foundation of Korea(NRF) funded by the Korea government(MSIT) (No. 2016M3C4A7952632), Industrial Strategic Technology Development Program(No. 10073144) funded by the Ministry of Trade, Industry \& Energy(MOTIE, Korea) \\

\bibliography{semeval2018}
\bibliographystyle{acl_natbib}

\appendix

\end{document}